\documentclass[letterpaper]{article}
\usepackage{aaai18}
\usepackage{times}
\usepackage{helvet}
\usepackage{courier}
\usepackage{url}
\usepackage{graphicx}
\usepackage{multirow}
\usepackage{amsmath,amssymb,amsthm}
\usepackage[noend]{algpseudocode}
\usepackage{algorithm}
\usepackage[inline]{enumitem}
\usepackage{array,makecell}

\usepackage{rotating,float}

\usepackage{booktabs}

\usepackage[capitalise,noabbrev]{cleveref}

\newcommand{\hidact}[2]{\phi^{{#2}}_{{#1}}}
\newcommand{\hidprop}[2]{{\psi^{{#2}}_{{#1}}}}
\newcommand{\inact}[2]{u^{{#2}}_{{#1}}}
\newcommand{\inprop}[2]{v^{{#2}}_{{#1}}}
\DeclareMathOperator{\getpred}{pred}
\DeclareMathOperator{\getop}{op}

\DeclareMathOperator{\pool}{pool}

\usepackage{color}

\begin{document}
\title{
  Action Schema Networks: Generalised Policies with Deep Learning
}

\author{
  Sam Toyer\textsuperscript{1}, Felipe Trevizan\textsuperscript{1,2}, Sylvie
  Thi\'ebaux\textsuperscript{1} {\rm and} Lexing Xie\textsuperscript{1,3}\\
  \textsuperscript{1} Research School of Computer Science, Australian National University
  \hspace{4pt}
  \textsuperscript{2} Data61, CSIRO\hspace{4pt}
  \textsuperscript{3} Data to Decisions CRC\\
  \texttt{first.last@anu.edu.au}
}

\maketitle

\begin{abstract}
In this paper, we introduce the Action Schema Network (ASNet): a neural network
architecture for learning generalised policies for probabilistic planning
problems.
By mimicking the relational structure of planning problems, ASNets are able to
adopt a weight sharing scheme which allows the network to be applied to any
problem from a given planning domain.
This allows the cost of training the network to be amortised over all problems
in that domain.
Further, we propose a training method which balances exploration and supervised
training on small problems to produce a policy which remains robust when
evaluated on larger problems.
In experiments, we show that ASNet's learning capability allows it to
significantly outperform traditional non-learning planners in several
challenging domains.
\end{abstract}

\section{Introduction}

Automated planning is the task of finding a sequence of actions which will
achieve a goal within a user-supplied model of an environment.
Over the past four decades, there has been a wealth of research into the use of
machine learning for automated planning~\cite{jimenez2012review}, motivated in
part by the belief that these two essential ingredients of
intelligence---planning and learning---ought to strengthen one
other~\cite{zimmerman2003learning}.
Nevertheless, the dominant paradigm among state-of-the-art classical and
probabilistic planners is still based on heuristic state space search.
The domain-independent heuristics used for this purpose are capable of
exploiting common structures in planning problems, but do not learn from
experience.
Top planners in both the deterministic and learning tracks of the International
Planning Competition often use machine learning to configure
portfolios~\cite{vallati2015ipc}, but only a small fraction of planners make
meaningful use of learning to produce domain-specific heuristics or control
knowledge~\cite{de2008learning}.
Planners which transfer knowledge between problems in a domain have been
similarly underrepresented in the probabilistic track of the competition.

\begin{figure}[ht!]
\includegraphics[width=\linewidth]{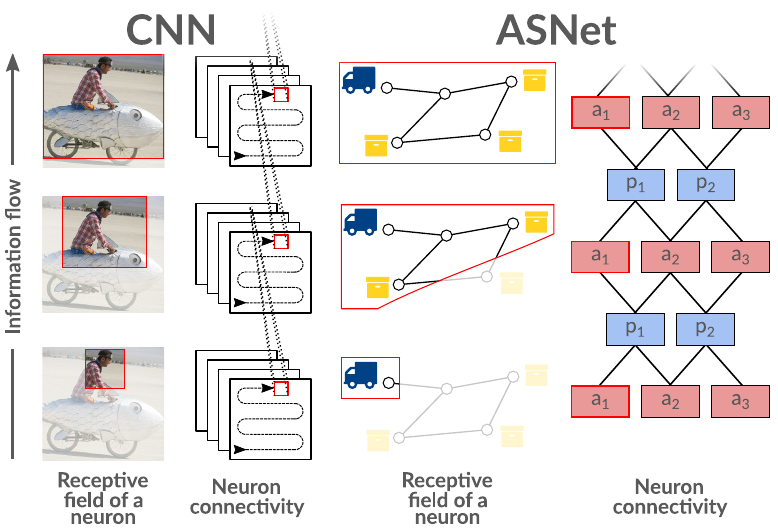}
\caption{
In a CNN, successive convolutions grow the receptive field of a neuron at higher
layers; analogously for ASNet, a neuron for a particular action $a$ or
proposition $p$ {\em sees} a larger portion of the current state at higher
layers.
}\label{fig:cnn-analogy}
\end{figure}

In parallel with developments in planning, we've seen a resurgence of interest
in neural nets, driven largely by their success at problems like image
recognition~\cite{krizhevsky2012imagenet} and learning to play video
games~\cite{mnih2013playing}.
This paper brings some gains of deep learning to planning by proposing a new
neural network architecture, the ASNet, which is specialised to the structure of
planning problems much as Convolutional Neural Networks (CNNs) are specialised
to the structure of images.
The basic idea is illustrated in \cref{fig:cnn-analogy}: rather than operating
on a virtual graph of pixels with edges defined by adjacency relationships,
an ASNet operates on a graph of actions and propositions (i.e. Boolean variables),
with edges defined by relations of the form ``action $a$ affects proposition
$p$'' or ``proposition $p$ influences the outcome of action $a$''.
This structure allows an ASNet to be trained on one problem from a given
planning domain and applied to other, different problems without re-training.

We make three new contributions.
\begin{enumerate*}[label={(\arabic*)}]
 \item A neural network architecture for probabilistic planning that
   automatically generalises to any problem from a given planning domain.
 \item A representation that allows weight sharing among actions modules
   belonging to the same action schema, and among proposition modules associated
   with the same predicate. This representation is augmented by input features
   from domain-independent planning heuristics.
 \item A training method that balances exploration and supervision from existing planners. 
\end{enumerate*}
In experiments, we show that this strategy is sufficient to learn effective
generalised policies.
Code and models for this work are available online.
\footnote{\texttt{https://github.com/qxcv/asnets}}
 
\section{Background}

\newcommand{\m}[1]{\ensuremath{\mathcal{#1}}}

This work considers probabilistic planning problems represented as Stochastic
Shortest Path problems (SSPs)~\cite{bertsekas96:neurodynamic}.
Formally, an SSP is a tuple $(\mathcal S, \mathcal A, \mathcal T, \mathcal C,
\mathcal G, s_0)$ where
$\mathcal S$ is a finite set of states,
$\mathcal A$ is a finite set of actions,
$\mathcal T\colon \mathcal S \times \mathcal A \times \mathcal S \to [0, 1]$ is
a transition function,
$\mathcal C\colon \mathcal S \times \mathcal A \to (0, \infty)$ is a cost
function,
$\mathcal G \subseteq \mathcal S$ is a set of goal states,
and $s_0$ is an initial state.
At each state $s$, an agent chooses an action $a$ from a set of enabled actions
$\m A(s) \subseteq \mathcal A$, incurring a cost of $\m C(s,a)$ and causing it
to transition into another state $s' \in \mathcal S$ with probability $\mathcal
T(s, a, s')$.

The solution of an SSP is a policy $\pi\colon \m{A} \times \m{S} \to [0, 1]$
such that $\pi(a\,|\,s)$ is the probability that action $a$ will be applied in
state $s$.
An optimal policy $\pi^*$ is any policy that minimises the total expected cost
of reaching $\mathcal{G}$ from $s_0$.
We do not assume that the goal is reachable with probability~1 from $s_0$ (i.e.
we allow problems with unavoidable dead ends),
and a fixed-cost penalty is incurred every time a dead end is
reached~\cite{mausam:kolobov:12}.

A \textit{factored SSP} is a compact representation of an SSP as a tuple
$(\mathcal{P}, \m{A}, s_0, s_\star, \m{C})$. 
$\mathcal{P}$ is a finite set of binary \textit{propositions}
and the state space $\mathcal{S}$ is the set of all binary strings of size
$|\mathcal{P}|$. Thus, a state $s$ is a value assignment to all the
propositions $p \in \mathcal{P}$.
A \textit{partial state} is a value assignment to a subset of propositions; a
partial state $s$ is consistent with a partial state $s'$ if the value
assignments of $s'$ are contained in $s$ ($s' \subseteq s$ for short).
The goal is represented by a partial state $s_\star$, and $\mathcal{G} = \{s
\in \m{S}|s_\star \subseteq s\}$.
Each action $a \in \m{A}$ consists in a precondition $\mathit{pre}_a$
represented by a partial state, a set of effects $\mathit{eff}_a$ each
represented by a partial state, and a probability distribution $\mathit{Pr}_a$
over effects in $\mathit{eff}_a$.\footnotemark{}
\footnotetext{
Factored SSPs sometimes support conditional effects and negative or disjunctive
preconditions and goals. 
We do not use these here to simplify notation. However, ASNet\ can easily be
extended to support these constructs.
}
The actions applicable in state $s$ are $\mathcal A(s) = \{a \in \m{A} \mid
\mathit{pre}_a \subseteq s\}$.
Moreover, $\mathcal{T}(s,a,s') = \sum_{e\in \mathit{eff}_a | s' = \mathit{res}(s,e)}
\mathit{Pr}_a(e)$ where 
$\mathit{res}(s,e) \in \m{S}$
is the result of changing the value of propositions of $s$ to make it consistent
with effect $e$.

A {\em lifted} SSP compactly represents a set of factored SSPs sharing the
same structure.
Formally, a lifted SSP is a tuple $(\m{F}, \mathbb{A}, \m{C})$ where $\m{F}$ is
a finite set of predicates, and $\mathbb{A}$ is a finite set of action schemas.
Each predicate, when grounded, i.e., instantiated by a tuple of names
representing objects, yields a factored SSP proposition. Similarly, each action
schema, instantiated by a tuple of names, yields a factored SSP action.
The Probabilistic Planning Domain Definition Language (PPDDL) is the standard
language to describe lifted and factored SSPs~\cite{younes2004ppddl1}.
PPDDL splits the description into a general {\em domain} and a specific {\em
problem}.
The domain gives the predicates \m{F}, action schemas $\mathbb{A}$ and cost
function \m{C} specifying a lifted SSP.
The problem additionally gives the set of objects \m{O}, initial state $s_0$ and
goal $s_\star$, describing a specific SSP whose propositions and actions are
obtained by grounding the domain predicates and action schemas using the objects
in \m{O}.
For instance the domain description might specify a predicate
$\operatorname{at}(?l)$ and an action schema
$\operatorname{walk}(?\mathrm{from}, ?\mathrm{to})$, while the problem
description might specify objects $\mathrm{home}$ and $\mathrm{work}$.
Grounding using these objects would produce propositions
$\operatorname{at}(\mathrm{home})$ and $\operatorname{at}(\mathrm{work})$, as
well as ground actions $\operatorname{walk}(\mathrm{work}, \mathrm{home})$ and
$\operatorname{walk}(\mathrm{home}, \mathrm{work})$.

Observe that different factored SSPs can be obtained by changing only the
problem part of the PPDDL description while reusing its domain.
In the next section, we show how to take advantage of action schema reuse to
learn policies that can then be applied to any factored SSP obtained by
instantiating the same domain.
 
\section{Action Schema Networks}

Neural networks are expensive to train, so we would like to amortise that cost
over many problems by learning a \textit{generalised policy} which can be
applied to any problem from a given domain.
ASNet proposes a novel, domain-specialised structure that uses the same set of
learnt weights $\theta$ regardless of the ``shape'' of the problem.
The use of such a weight sharing scheme is key to ASNet's ability to generalise
to different problems drawn from the same domain, even when those problems have
different goals or different numbers of actions and propositions.

\subsection{Network structure}\label{sec:net-struct}

At a high level, an ASNet is composed of alternating action layers and
proposition layers, where action layers are composed of a single \textit{action
module} for each ground action, and proposition layers likewise are composed of
a single \textit{proposition module} for each ground proposition; this choice of
structure was inspired by the alternating action and proposition layers of
Graphplan~\cite{blum1997fast}.
In the same way that hidden units in one layer of a CNN connect only to nearby
hidden units in the next layer, action modules in one layer of an ASNet connect
only to directly related proposition modules in the next layer, and vice versa.
The last layer of an ASNet is always an action layer with each module defining
an action selection probability, thus allowing the ASNet to scale to problems
with different numbers of actions. For simplicity, we also assume that the first
(input) layer is always an action layer.

\textbf{Action module details.} Consider an action module for \mbox{$a \in \cal A$}
in the $l$th action layer. The module takes as input a feature vector
$\inact{a}{l}$, and produces a new hidden representation
\[
  \hidact{a}{l} = f(W_a^l \cdot \inact{a}{l} + b_a^l)~,
\]
where $W_a^l \in \mathbb R^{d_h \times d_a^l}$ is a learnt weight matrix for the
module, $b_a^l \in \mathbb R^{d_h}$ is a learnt bias vector, $f(\cdot)$ is a
nonlinearity (e.g. $\tanh$, sigmoid, or ReLU), $d_h$ is a (fixed) intermediate
representation size, and $d_a^l$ is the size of the inputs to the action module.
The feature vector $\inact{a}{l}$, which serves as input to the action module,
is constructed by enumerating the propositions $p_1, p_2, \ldots, p_M$ which are
\textit{related} to the action $a$, and then concatenating their hidden
representations.
Formally, we say that a proposition $p \in \m{P}$ is related to an action $a \in
\m{A}$, denoted $R(a, p)$, if $p$ appears in $\mathit{pre}_a$ or in an effect
$e$ where $\mathit{Pr}_a(e) > 0$.
Concatenation of representations for the related propositions produces a vector
\[
  \inact{a}{l} = \begin{bmatrix}
    \hidprop{1}{l-1}^T  & \cdots & \hidprop{M}{l-1}^T
  \end{bmatrix}^T~,
\]
where $\hidprop{j}{l-1}$ is the hidden representation produced by the
proposition module for proposition $p_j \in \m{P}$ in the preceding proposition
layer.
Each of these constituent hidden representations has dimension $d_h$, so
$\inact{a}{l}$ has dimension $d_a^l = d_h \cdot M$.

Our notion of propositional relatedness ensures that, if ground actions $a_1$
and $a_2$ in a problem are 
instances of the same action schema in a PPDDL
domain, then their inputs $\inact{1}{l}$ and $\inact{2}{l}$ will have the same
``structure''. To see why, note that we can determine which propositions are
related to a given ground action $a$ by retrieving the corresponding action
schema, enumerating the predicates which appear in
the precondition or the effects of the
action schema, then instantiating those predicates with the same parameters used
to instantiate $a$. If we apply this procedure to $a_1$ and
$a_2$, we will obtain lists of related propositions $p_1, p_2, \ldots, p_M$ and
$q_1, q_2, \ldots, q_M$, respectively, where $p_j$ and $q_j$ are propositions
with the same predicate which appear in the same position in the definitions of
$a_1$ and $a_2$ (i.e. the same location in the precondition, or the same
position in an effect).

Such structural similarity is key to ASNet's generalisation abilities.
At each layer $l$, and for each pair of ground actions $c$ and $d$ instantiated
from the same action schema $s$, we use the same weight matrix $W^l_s$ and bias
vector $b^l_s$---that is, we have $W_c^l = W_d^l = W^l_s$ and $b_c^l = b_d^l =
b^l_s$.
Hence, modules for actions which appear in the same layer and correspond to the
same action schema will use the same weights, but modules which appear in
different layers or which correspond to different schemas will learn different
weights.
Although different problems instantiated from the same PPDDL domain may have
different numbers of ground actions, those ground actions will still be derived
from the same, fixed set of schemas in the domain, so we can apply the same set
of action module weights to any problem from the domain.

The first and last layers of an ASNet consist of action modules, but their
construction is subtly different:
\begin{enumerate}
  \item The output of a module for action $a$ in the final layer is a single number
    $\pi^\theta(a \mid s)$ representing the probability of selecting action $a$
    in the current state $s$ under the learnt policy $\pi^\theta$, rather than a
    vector-valued hidden representation. To guarantee that disabled actions are
    never selected, and ensure that action probabilities are normalised to 1, we
    pass these outputs through a \textit{masked softmax} activation which ensures
    that $\pi^\theta(a \mid s) = 0$ if $a \notin \m{A}(s)$. During training, we
    sample actions from $\pi^\theta(a \mid s)$. During evaluation, we select the
    action with the highest probability.

  \item Action modules in the first layer of a ASNet are passed an input
    vector composed of features derived from the current state, rather than
    hidden representations for related propositions. Specifically, modules in
    the first layer are given a binary vector indicating the truth values
    of related propositions, and whether those propositions appear in the goal.
    In practice, it is helpful to concatenate these
    propositional features with heuristic features, as described in
    Section~\ref{ssec:heuristic}.
\end{enumerate}

\textbf{Proposition module details.} Proposition modules only appear in the
intermediate layers of an ASNet, but are otherwise similar to action
modules.
Specifically, a proposition module for proposition $p \in \m{P}$ in the $l$th
proposition layer of the network will compute a hidden representation
\[
  \hidprop{p}{l} = f(W_p^l \cdot \inprop{p}{l} + b_p^l)~,
\]
where $\inprop{p}{l}$ is a feature vector, $f$ is the same nonlinearity used
before, and $W_p^l \in \mathbb R^{d_h \times d_p^l}$ and $b_p^l \in \mathbb
R^{d_h}$ are learnt weights and biases for the module.

To construct the input $\inprop{p}{l}$, we first find the predicate $\getpred(p)
\in \m{F}$ for proposition $p \in \m{P}$, then enumerate all action schemas
$A_1, \ldots, A_L \in \mathbb{A}$ which reference $\getpred(p)$ in a
precondition or effect.
We can define a feature vector
\[
  \inprop{p}{l} = \begin{bmatrix}
    \pool(\{{\hidact{a}{l}}^T \mid \mathcal \getop(a) = A_1 \land R(a, p)\})\\
    \vdots\\
    \pool(\{{\hidact{a}{l}}^T \mid \mathcal \getop(a) = A_L \land R(a, p)\})\\
  \end{bmatrix}~,
\]
where $\getop(a) \in \mathbb{A}$ denotes the action schema for ground
action $a$, and $\pool$ is a pooling function that combines several
$d_h$-dimensional feature vectors into a single $d_h$-dimensional one.
Hence, when all pooled vectors are concatenated, the dimensionality $d_p^l$ of
$\inprop{p}{l}$ becomes $d_h \cdot L$.
In this paper, we assume that $\pool$ performs max pooling (i.e. keeps only the
largest input).
If a proposition module had to pool over the outputs of many action modules,
such pooling could potentially obscure useful information.
While the issue could be overcome with a more sophisticated pooling mechanism
(like neural attention), we did not find that max pooling posed a major problem
in the experiments in \cref{sec:expts}, even on large Probabilistic Blocks World
instances where some proposition modules must pool over thousands of inputs.

Pooling operations are essential to ensure that proposition modules
corresponding to the same predicate have the same structure.
Unlike action modules corresponding to the same action schema, proposition
modules corresponding to the same predicate may have a different number of
inputs depending on the initial state and number of objects in a problem, so it
does not suffice to concatenate inputs.
As an example, consider a single-vehicle logistics problem where the location of
the vehicle is tracked with propositions
of the form $\operatorname{at}(\iota)$, and the vehicle may be moved with
actions of the form $\operatorname{move}(\iota_{\text{from}},
\iota_{\text{to}})$. A location $\iota_1$ with one incoming road and no outgoing
roads will have only one related $\operatorname{move}$ action, but a location
$\iota_2$ with two incoming roads and no outgoing roads will have two related
$\operatorname{move}$ actions, one for each road.
This problem is not unique to planning: a similar trick is employed in network
architectures for graphs where vertices can have varying
in-degree~\cite{jain2016structural,kearnes2016molecular}.

As with the action modules, we share weights between proposition modules for
propositions corresponding to the same predicate.
Specifically, at proposition layer $l$, and for propositions $q$ and $r$ with
$\getpred(q) = \getpred(r)$, we tie the corresponding weights $W_q^l = W_r^l$
and $b_q^l = b_r^l$.
Together with the weight sharing scheme for action modules, this enables us to
learn a single set of weights
\[
\begin{split}
  \theta =
  &\{W_a^l, b_a^l \mid 1 \leq l \leq n + 1, a \in \mathbb A\}\\
  \cup
  &\{W_p^l, b_p^l \mid 1 \leq l \leq n, p \in \m{F}\}
\end{split}
\]
for an $n$-layer model which can be applied to any problem in a given PPDDL domain.

\subsection{Heuristic features for expressiveness}
\label{ssec:heuristic}

One limitation of the ASNet is the fixed receptive field of the network; in
other words, the longest chain of related actions and propositions which it can
reason about.
For instance, suppose we have $I$ locations $\iota_1, \ldots, \iota_I$ arranged
in a line in our previous logistics example.
The agent can move from $\iota_{k-1}$ to $\iota_k$ (for $k\!=\!2,\ldots,I$) with
the $\operatorname{move}(\iota_{k-1}, \iota_{k})$ action, which makes
$\operatorname{at}(\iota_{k-1})$ false and $\operatorname{at}(\iota_{k})$ true.
The propositions $\operatorname{at}(\iota_{1})$ and $\operatorname{at}(\iota_I)$
will thus be related only by a chain of $\operatorname{move}$ actions of length
$I-1$; hence, a proposition module in the $l$th proposition layer will only be
affected by $\operatorname{at}$ propositions for locations at most $l+1$ moves
away.
Deeper networks can reason about longer chains of actions, but that an ASNet's
(fixed) depth necessarily limits its reasoning power when chains of actions can
be \textit{arbitrarily} long.

We compensate for this receptive field limitation by supplying the network with
features obtained using domain-independent planning heuristics.
In this paper, we derive these features from disjunctive action landmarks
produced by LM-cut~\cite{helmert2009landmarks}, but features derived from
different heuristics could be employed in the same way.
A disjunctive action landmark is a set of actions in which at least one action
must be applied along any optimal path to the goal in a deterministic,
delete-relaxed version of the planning problem. These landmarks do not
necessarily capture all useful actions, but in practice we find that providing
information about these landmarks is often sufficient to compensate for network
depth limitations.

In this paper, a module for action $a$ in the first network layer
is given a feature vector
\[
  \inact{a}{1} = \begin{bmatrix}c^T & v^T & g^T\end{bmatrix}^T.
\]
$c \in \mathbb \{0, 1\}^3$ indicates whether $a_i$ is the sole action in at
least one LM-cut landmark ($c_1 = 1$), an action in a landmark of two or more
actions ($c_2 = 1$), or does not appear in a landmark ($c_3 = 1$). $v \in \{0,
1\}^M$ represents the $M$ related propositions: $v_j$ is 1 iff $p_j$ is
currently true. $g \in \{0, 1\}^M$ encodes related portions of the goal state,
and $g_j$ is 1 iff $p_j$ is true in the partial state $s_\star$ defining the
goal.
 
\section{Training with exploration and supervision}

We learn the ASNet weights $\theta$ by choosing a set of small training problems
$P_{\text{train}}$, then alternating between guided exploration to build up a
state memory $\mathcal M$, and supervised learning to ensure that the network
chooses good actions for the states in $\mathcal M$.
\cref{alg:training} describes a single epoch of exploration and
supervised learning. We repeatedly apply this procedure until performance on
$P_{\text{train}}$ ceases to improve, or until a fixed time limit is reached.
Note that this strategy is only intended to learn the \textit{weights} of an
ASNet---module connectivity is not learnt, but rather obtained from a grounded
representation using the notion of relatedness which we described earlier.

\begin{algorithm}[t]
\begin{algorithmic}[1]
\Procedure{ASNet-Train-Epoch}{$\theta$, $\mathcal M$}
  \For{$i=1, \ldots, T_{\text{explore}}$}
    \Comment{Exploration}
    \ForAll{$\zeta \in P_{\text{train}}$}
      \State{$s_0, \ldots, s_N \gets \textsc{Run-Pol}(s_0(\zeta), \pi^\theta)$}
      \State{$\mathcal M \gets \mathcal M \cup \{s_0, \ldots, s_N\}$}
      \For{$j = 0, \ldots, N$}
        \State{$s_j^*, \ldots, s_{M}^* \gets \textsc{Pol-Envelope}(s_j, \pi^*)$}
        \State{$\mathcal M \gets \mathcal M \cup \{s_j^*, \ldots, s_{M}^*\}$}
      \EndFor{}
    \EndFor{}
  \EndFor{}
  \For{$i = 1, \ldots, T_{\text{train}}$}
  \Comment{Learning}
    \State{$\mathcal B \gets \textsc{Sample-Minibatch}(\mathcal M)$}
    \State{Update $\theta$ using $\frac{d\mathcal L_\theta(\mathcal B)}{d\theta}$ (Equation~\ref{eqn:batch-objective})}
  \EndFor{}
\EndProcedure{}
\end{algorithmic}
\caption{Updating ASNet weights $\theta$ using state memory $\mathcal M$ and
  training problem set $P_{\text{train}}$}
\label{alg:training}
\end{algorithm}

In the exploration phase of each training epoch, we repeatedly run the ASNet
policy $\pi^\theta$ from the initial state of each problem $\zeta \in
P_{\text{train}}$, collecting $N+1$ states $s_{0}, \ldots, s_N$ visited
along each of the sampled trajectories.
Each such trajectory terminates when it reaches a goal, exceeds a fixed limit
$L$ on length, or reaches a dead end.
In addition, for each visited state $s_{j}$, we compute an optimal policy
$\pi^*$ rooted at $s_j$, then produce a set of states $s^*_{j}, \ldots, s^*_{M}$
which constitute $\pi^*$'s policy envelope---that is, the states which $\pi^*$
visits with nonzero probability.
Both the trajectories drawn from the ASNet policy $\pi^\theta$ and policy
envelopes for the optimal policy $\pi^*$ are added to the state memory $\mathcal
M$.
Saving states which can be visited under an optimal policy ensures that
$\mathcal M$ always contains states along promising trajectories reachable from
$s_0$.
On the other hand, saving trajectories from the exploration policy ensures that
ASNet will be able to improve on the states which it visits most often, even if
they are not on an optimal goal trajectory.

In the training phase, small subsets of the states in $\mathcal M$ are
repeatedly sampled at random to produce minibatches for training ASNet.
The objective to be minimised for each minibatch $\mathcal B$ is the
cross-entropy classification loss
\begin{equation}\label{eqn:batch-objective}
  \begin{split}
  \mathcal L_\theta(\mathcal B) = \sum_{s \in \mathcal B}\sum_{a \in A} \big[
    & (1 - y_{s,a}) \cdot \log (1 - \pi^\theta(a \mid s))\\
    &+ y_{s, a} \cdot \log \pi^\theta(a \mid s)
  \big]~.
  \end{split}
\end{equation}
The label $y_{s,a}$ is 1 if the expected cost of choosing action $a$ and then
following an optimal policy thereafter is minimal among all enabled actions;
otherwise, $y_{s, a} = 0$.
This encourages the network to imitate an optimal policy.
For each sampled batch $\mathcal B$, we compute the gradient $\frac{d\mathcal
L_\theta(\mathcal B)}{d\theta}$ and use it to update the weights $\theta$ in a
direction which decreases
$\mathcal L_\theta(\mathcal B)$ with Adam~\cite{kingma2014adam}.

The cost of computing an optimal policy during supervised learning is often
non-trival. It is natural to ask whether it is more efficient to train ASNets
using unguided policy gradient reinforcement learning, as FPG
does~\cite{buffet2009factored}. Unfortunately, we found that policy gradient RL
was too noisy and inefficient to train deep networks on nontrivial problems; in
practice, the cost of computing an optimal policy for small training problems
more than pays for itself by enabling us to use sample-efficient supervised
learning instead of reinforcement learning. In the experiments, we investigate the
question of whether suboptimal policies are still sufficient for supervised
training of ASNets.

Past work on generalised policy learning has employed learnt policies as control
knowledge for search algorithms, in part because doing so can compensate for
flaws in the policy. For example,
\citeauthor{yoon2007using}~(\citeyear{yoon2007using}) suggest employing policy
rollout or limited discrepancy search to avoid the occasional bad action
recommended by a policy. While we could use an ASNet similarly, we are
more interested in its ability to learn a reliable policy on its own. Hence,
during evaluation, we always choose the action which maximises $\pi^\theta(a
\mid s)$. As noted above, this is different from the exploration process
employed during training, where we instead sample from $\pi^\theta(a \mid s)$. 
\section{Experiments and discussion}\label{sec:expts}

In this section, we compare ASNet against state-of-the-art planners on three
planning domains.

\subsection{Experimental setup}\label{sec:methodology}

We compare ASNet against three heuristic-search-based probabilistic planners:
LRTDP~\cite{bonet:geffner:03}, ILAO*~\cite{hansen:zilberstein:01} and
SSiPP~\cite{trevizan14:depth}.
Two domain-independent heuristics are considered for each of the three
planners---LM-cut (admissible) and the additive heuristic $h^{\text{add}}$
(inadmissible)~\cite{teichteil11:heuristic}---resulting in 6 baselines.
During evaluation, we enforce a 9000s time cutoff for all the baselines and
ASNets, as well as a 10Gb memory cutoff.

Since LRTDP and ILAO* are optimal planners, we execute them until convergence
($\epsilon\!=\!10^{-4}$) for each problem using 30 different random seeds.
Notice that, for $h^{\text{add}}$, LRTDP and ILAO* might converge to a
suboptimal solution.
If an execution of LRTDP or ILAO* does not converge before the given time/memory
cutoff, we consider the planner as having failed to reach the goal.
SSiPP is used as a replanner and, for each problem, it is \textit{trained} until
60s before the time cutoff and then evaluated; this procedure is repeated 30
times for each problem using different random seeds.
The training phase of SSiPP consists in simulating a trajectory from $s_0$ and,
during this process, SSiPP improves its lower bound on the optimal solution.
If 100 consecutive trajectories reach the goal during training, then SSiPP is
evaluated regardless of the training time left.
For the 6 baselines, we report the average running time per problem.

For each domain, we train a single ASNet, then evaluate it on each problem 30
times with different random seeds.
The hyperparmeters for each ASNet were kept fixed across domains: three action
layers and two proposition layers in each network, a hidden representation size
of 16 for each internal action and proposition module, and an
ELU~\cite{clevert2015fast} as the nonlinearity $f$.
The optimiser was configured with a learning rate of 0.0005 and a batch size of
128, and a hard limit of two hours (7200s) was placed on training.
We also applied $\ell_2$ regularisation with a coefficient of 0.001 on all
weights, and dropout on the outputs of each layer except the last with $p =
0.25$.
Each epoch of training alternated between 25 rounds of exploration shared
equally among all training problems, and 300 batches of network optimisation
(i.e. $T_{\text{explore}} = 25 / |P_{\text{train}}|$ and $T_{\text{train}} =
300$).
Sampled trajectory lengths are $L = 300$ for both training and evaluation.
LRTDP with the LM-cut heuristic is used for computing the optimal policies
during training, with a dead-end penalty of 500.
We also repeated this procedure for LRTDP using $h^{\text{add}}$ (inadmissible
heuristic) to compare the effects of using optimal and suboptimal policies for
training.
Further, we report how well ASNet performs when it is guided by
$h^{\text{add}}$, but \textit{not} given the LM-cut-derived heuristic features
described in \cref{ssec:heuristic}.
For the ASNets, we report the average \textit{training time plus time to solve
the problem} to highlight when it pays off to spend the one-off cost of
training an ASNet for a domain.

All ASNets were trained and evaluated on a virtual machine equipped with 62GB of
memory and an x86-64 processor clocked at 2.3GHz.
For training and evaluation, each ASNet was restricted to use a single,
dedicated processor core, but resources were otherwise shared.
The baseline planners were run in a cluster of x86-64 processors clocked at
2.6GHz and each planner again used only a single core.

\begin{figure*}
  \includegraphics{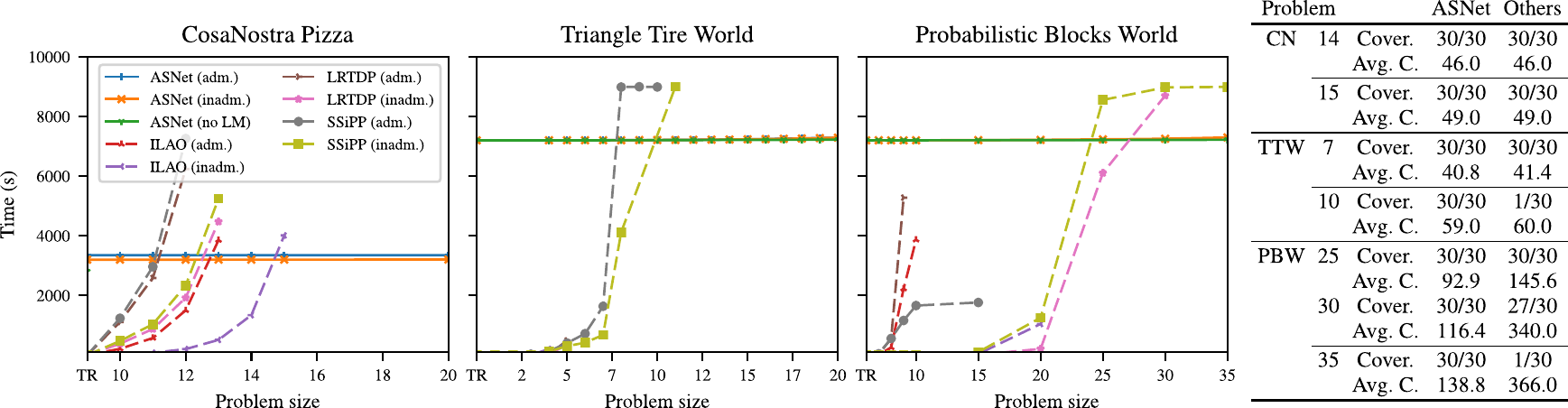}

  \caption{
    Comparison of planner running times on the evaluation domains. TR refers to
    the time used for training (zero for baselines). ASNet runs with (adm.) used
    optimal policies for training while (inadm.) used potentially suboptimal
    policies, and runs with (no LM) did not use heuristic input features.
    The table at right shows, for selected problems, the coverage and average
    solution cost for the best ASNet and baseline.
    We use TTW for Triangle Tire World, CN for CosaNostra Pizza, and PBW for
    Probabilistic Blocks World.
    In PBW, running times are averaged over the three problems of each size.
    In TTW and PBW, ASNet (no LM) occludes ASNet (inadm.).
    ASNet (adm.) is also occluded in TTW, but is absent entirely from PBW as the
    optimal planner used to generate training data could not solve all training
    problems in time.
  }
  \label{fig:ctimes-ttw}
\end{figure*}

\subsection{Domains}\label{sec:domains}

We evaluate ASNets and the baselines on the following probabilistic planning
domains:

\textbf{CosaNostra Pizza:} as a Deliverator for CosaNostra Pizza, your job is to
safely transport pizza from a shop to a waiting customer, then return to the
shop.
There is a series of toll booths between you and the customer: at each booth,
you can either spend a time step paying the operator, or save a step by driving
through without paying.
However, if you don't pay, the (angry) operator will try to drop a boom on your
car when you pass through their booth on the way back to the shop, crushing the
car with 50\% probability.
The optimal policy is to pay operators when travelling to the customer to ensure
a safe return, but not pay on the return trip as you will not revisit the booth.
Problem size is the number of toll booths between the shop and the customer.
ASNets are trained on sizes 1-5, and tested on sizes 6+.

\textbf{Probabilistic Blocks World} is an extension of the well-known
deterministic blocks world domain in which a robotic arm has to move blocks on
a table into a goal configuration.
The actions to pick up a block or to put a block on top of another fail with
probability 0.25; failure causes the target block to drop onto the table,
meaning that it must be picked up and placed again.
We randomly generate three different problems for each number of blocks
considered during testing.
ASNet is trained on five randomly generated problems of each size from
5--9, for 25 training problems total.

\textbf{Triangle Tire World~\cite{little2007probabilistic}:} each problem
consists of a set of locations arranged in a triangle, with connections between
adjacent locations.
The objective is to move a vehicle from one corner of the triangle to another.
However, each move has a 50\% chance of producing a flat tire, which must be
replaced at the next visited location.
The vehicle thus requires a sequence of moves between locations where
replacement tires are available.
Tires are arranged such that the most reliable policy is one which travels the
\textit{longest} path to the goal, along the outside edge of the triangle.
This task can be made more challenging by scaling up the number of locations.
Per \citeauthor{little2007probabilistic}~(\citeyear{little2007probabilistic}), a
problem of size $n$ has $(n+1)(2n+1)$ locations.
We use sizes 1-3 for training, and test with sizes from 4 onward.

\subsection{Results}\label{ssec:results}

\cref{fig:ctimes-ttw} shows the time taken to train and evaluate ASNet using
optimal (adm.) and suboptimal (inadm.) policies as training data.
In addition, it shows coverage (proportion of runs which reached the goal) and
average solution cost when the goal is reached for selected problems for the
best ASNet and best baseline.
The following is a summary of our results:

\textbf{When is it worth using ASNet?} All ASNets obtained 30 out of 30
coverage for all Triangle Tire World problems, and the ASNets with heuristic
input features similarly obtained perfect coverage on CosaNostra.
In contrast, the baselines failed to scale up to the larger problems.
This shows that ASNet is well-suited to problems where local knowledge of the
environment can help to avoid common traps, for instance:
in CosaNostra, the agent must learn to pay toll booth operators when carrying a
pizza and not pay otherwise;
and in Triangle Tire World, the agent must learn to sense and follow the outer
edge of the triangle.
Not only could ASNets learn these tricks, but the average solution cost obtained
by ASNets for CosaNostra and Triangle Tire World was close to that of the
optimal baselines (when they converged), suggesting that the optimal solution
was found.

Probabilistic Blocks World is more challenging as there is no single pattern
that can solve all problems.
Even for the deterministic version of Blocks World, a generalised policy
requires the planner to learn a recursive property for whether each block is in
a goal position~\cite{slaney2001blocks}.
The ASNet appears to have successfully learnt to do this when trained by a
suboptimal teacher and given landmarks as input, and surpassed all baselines in
coverage (reaching the goal on 30/30 runs on each instance).
Moreover, the average solution cost of ASNet (inadm.) is similar to the optimal
baselines (when they converge) and up to 3.7 times less than SSiPP (inadm.), the
baseline with the best coverage.
The ASNet (inadm.) policy typically obtained a mean solution cost somewhere
between the US and GN1 strategies presented by \citeauthor{slaney2001blocks}: it
is suboptimal, but still better than unstacking and rebuilding all towers from
scratch.
Note that the ASNet could not obtain a policy within the allotted time when
trained by an optimal teacher.

\textbf{Are the heuristic features necessary?}
In some cases, ASNet's performance can be improved by omitting (expensive)
LM-cut heuristic input features.
For instance, in Triangle Tire World, ASNet (inadm.) took 2.4x as much time as
ASNet (no LM) to solve problems of size 15, and 4.3x as much time to solve
problems of size 20, despite executing policies of near-identical average cost.
Notice that this difference cannot be seen in \cref{fig:ctimes-ttw} because the
training time (TR) is much larger than the time to solve a test instance.

Interestingly, ASNet (no LM) was able to obtain 100\% coverage on the
Probabilistic Blocks World problems in \cref{fig:ctimes-ttw}, despite not
receiving landmark inputs.
To gain stronger assurance that it had learnt a robust policy, we tested on 10
more instances with 10, 15, 20, 25, 30 and 35 blocks (60 more instances total).
ASNet (no LM) could not solve all the additional test instances.
In contrast, ASNet (inadm.)---which \textit{was} given landmarks as
input---reliably solved all test problems in the extended set, thus showing that
heuristic inputs are necessary to express essential recursive properties like
whether a block is in its goal position.

Heuristic inputs also appear to be necessary in CosaNostra, where ASNet (no LM)
could not achieve full coverage on the test set.
We suspect that this is because an ASNet without heuristic inputs cannot
determine which direction leads to the pizza shop and which direction leads to
the customer when it is in the middle of a long chain of toll booths.

\textbf{How do suboptimal training policies affect ASNet?} Our results
suggest that use of a suboptimal policies is sufficient to train ASNet, as
demonstrated in all three domains.
Intuitively, the use of suboptimal policies for training ought to be beneficial
because the time that would have been spent computing an optimal policy can
instead be used for more epochs of exploration and supervised learning.
This is somewhat evident in CosaNostra---where a suboptimal training policy
allows for slightly faster convergence---but it is more clear in Probabilistic
Blocks World, where the ASNet can only converge within our chosen time limit
with the inadmissible policy.
While training on fewer problems allowed the network to converge within the time
limit, it did not yield as robust a policy, suggesting that the use of a
suboptimal teacher is sometimes a necessity.

\textbf{Is ASNet performing fixed-depth lookahead search?}
No.
This can be seen by comparing SSiPP and ASNet.
SSiPP solves fixed-depth sub-problems (a generalization of lookahead for SSPs)
and is unable to scale up as well as ASNets when using an equivalent depth
parametrisation.
Triangle Tire World is particularly interesting because SSiPP can outperform
other baselines by quickly finding dead ends and avoiding them.
However, unlike an ASNet, SSiPP is unable to generalize the solution of one
sub-problem to the next and needs to solve all of them from scratch.
 
\section{Related work}

Generalised policies are a topic of interest in
planning~\cite{zimmerman2003learning,jimenez2012review,hu2011generalized}.
The earliest work in this area expressed policies as decision
lists~\cite{khardon1999learning}, but these were insufficiently expressive to
directly capture recursive properties, and thus required user-defined
\textit{support predicates}.
Later planners partially lifted this restriction by expressing learnt rules with
\textit{concept language} or \textit{taxonomic syntax}, which can capture such
properties
directly~\cite{martin2000learning,yoon2002inductive,yoon2004learning}.
Other work employed features from domain-independent heuristics to capture
recursive properties~\cite{de2011scaling,yoon2006learning}, just as we do with
LM-cut landmarks.
\citeauthor{srivastava2011directed}~(\citeyear{srivastava2011directed}) have
also proposed a substantially different generalised planning strategy that
provides strong guarantees on plan termination and goal attainment, albeit only
for a restricted class of deterministic problems.
Unlike the decision lists~\cite{yoon2002inductive,yoon2004learning} and
relational decision trees~\cite{de2011scaling} employed in past work, our
model's input features are fixed before training, so we do not fall prey to
the \textit{rule utility problem}~\cite{zimmerman2003learning}.
Further, our model can be trained to minimise any differentiable loss, and could
be modified to use policy gradient reinforcement learning without changing the
model.
While our approach cannot give the same theoretical guarantees as
\citeauthor{srivastava2011directed}, we are able to handle a more general class
of problems with less domain-specific information.

Neural networks have been used to learn policies for probabilistic planning
problems.
The Factored Policy Gradient (FPG) planner trains a multi-layer perceptron with
reinforcement learning to solve a factored MDP~\cite{buffet2009factored}, but it
cannot generalise across problems and must thus be trained anew on each
evaluation problem.
Concurrent with this work,
\citeauthor{groshev2017learning}~(\citeyear{groshev2017learning}) propose
generalising ``reactive'' policies and heuristics by applying a CNN to a 2D
visual representation of the problem, and demonstrate an effective learnt
heuristic for Sokoban.
However, their approach requires the user to define an appropriate visual
encoding of states, whereas ASNets are able to work directly from a PPDDL
description.

The integration of planning and neural networks has also been investigated in
the context of deep reinforcement learning.
For instance, Value Iteration Networks~\cite{tamar2016value,niu2017generalized}
(VINs) learn to formulate and solve a probabilistic planning problem within a
larger deep neural network.
A VIN's internal model can allow it to learn more robust policies than would be
possible with ordinary feedforward neural networks.
In contrast to VINs, ASNets are intended to learn reactive policies for known
planning problems, and operate on factored problem representations instead of
(exponentially larger) explicit representations like those used by VINs.

In a similar vein, \citeauthor{kansky2017schema} present a model-based RL
technique known as schema networks~\cite{kansky2017schema}.
A schema network can learn a transition model for an environment which has been
decomposed into entities, but where those entities' interactions are initially
unknown.
The entity--relation structure of schema networks is reminiscent of the
action--proposition structure of an ASNet; however, the relations between ASNet
modules are obtained through grounding, whereas schema networks learn which
entities are related from scratch.
As with VINs, schema networks tend to yield agents which generalise well across
a class of similar environments.
However, unlike VINs and ASNets---which both learn policies directly---schema
networks only learn a model of an environment, and planning on that model must
be performed separately.

Extension of convolutional networks to other graph structures has received
significant attention recently, as such networks often have helpful invariances
(e.g. invariance to the order in which nodes and edges are given to the
network) and fewer parameters to learn than fully connected networks.
Applications include reasoning about spatio-temporal relationships between
variable numbers of entities~\cite{jain2016structural}, molecular
fingerprinting~\cite{kearnes2016molecular}, visual question
answering~\cite{teney2016graph}, and reasoning about knowledge
graphs~\cite{kipf2016semi}.
To the best of our knowledge, this paper is the first such technique that
successfully solves factored representations of automated planning problems. 
\section{Conclusion}

We have introduced the ASNet, a neural network architecture which is able to
learn generalised policies for probabilistic planning problems.
In much the same way that CNNs can generalise to images of arbitrary size by
performing only repeated local operations, an ASNet can generalise to different
problems from the same domain by performing only convolution-like operations on
representations of actions or propositions which are \textit{related} to one
another.
In problems where some propositions are only related by long chains of actions,
ASNet's modelling capacity is limited by its depth, but it is possible to avoid
this limitation by supplying the network with heuristic input features, thereby
allowing the network to solve a range of problems.

While we have only considered supervised learning of generalised policies, the
ASNet architecture could in principle be used to learn heuristics or embeddings,
or be trained with reinforcement learning.
ASNet only requires a model of which actions affect which portion of a state, so
it could also be used in other settings beyond SSPs, such as MDPs with
Imprecise Probabilities (MDPIPs)~\cite{white94:mdpitp} and MDPs with Set-Valued
Transitions (MDPSTs)~\cite{trevizan2007planning}.
We hope that future work will be able to explore these alternatives and use
ASNets to further enrich planning with the capabilities of deep learning.
 
{\small
\bibliography{citations,fwt}
\bibliographystyle{aaai}
}

\appendix
\section{Supplementary material}

\subsection{Monster experiments}

To illustrate when LM-cut flags are \textit{not} sufficient, we created a simple
domain called Monster, in which the agent must choose between two $n$-step paths
to reach the goal. This domain domain uses the same $\operatorname{at}(\iota)$
predicate and $\operatorname{move}(\iota_1, \iota_2)$ operators from the running
logistics example in the main paper. However, at the beginning of each episode,
a monster is randomly placed at the final location along one of the two paths,
and it has a 99\% chance of attacking the agent if the agent moves to the its
location. Since there is still a 1\% chance of not attacking the agent, an
all-outcome determinisation cannot indicate which path has the monster on it,
and so the agent must look ahead at least $n$ steps to be safe. Otherwise, if
the agent chooses at random, there is a 50\% chance that they will choose the
wrong path, and subsequently hit a dead end with high probability.

We perform an experiment on this domain in which we train ASNets with increasing
depth on problems with paths from length 1-5. We then test on those same
problems to determine what the agent was able to learn. Table~\ref{tab:monster}
shows the full results. As expected, the only runs with full coverage are those
where the ASNet has sufficient depth to see the monster; all others force the
ASNet to choose arbitrarily.

\begin{table}[H]
  \begin{center}
  \begin{tabular}{|c|ccccc|}
    \hline
    \multirow{2}{*}{\makecell{Proposition\\layers}} & \multicolumn{5}{c|}{Path length}\\
    & 1 & 2 & 3 & 4 & 5\\
    \hline
    1 & 30/30 & 14/30 & 14/30 & 14/30 & 14/30\\
    2 & 30/30 & 30/30 & 14/30 & 14/30 & 14/30\\
    3 & 30/30 & 30/30 & 30/30 & 14/30 & 14/30\\
    4 & 30/30 & 30/30 & 30/30 & 30/30 & 14/30\\
    \hline
  \end{tabular}
  \end{center}
  \caption{Coverage (out of 30) for Monster problem with different layer counts.}
  \label{tab:monster}
\end{table}

\section{Coverage and cost for probabilistic experiments}

To complement the time figures and basic overview of coverage given in the main
paper, \cref{tab:prob-res-ttw}, \cref{tab:prob-res-pbw}, \cref{tab:prob-res-cn}
and show coverage and solution cost for the evaluated probabilistic problems.

\begin{sidewaystable*}
\begin{center}
\begin{scriptsize}
\begin{tabular}{|l|ccc|cc|cc|cc|}
\hline
\multirow{2}{*}{Problem}
  & \multicolumn{3}{c|}{ASNet}
  & \multicolumn{2}{c|}{ILAO}
  & \multicolumn{2}{c|}{LRTDP}
  & \multicolumn{2}{c|}{SSiPP}
  \\
  & {LM-cut} & {$h^{\text{add}}$} & {$h^{\text{add}}$, no LM}
  & {LM-cut} & {$h^{\text{add}}$}
  & {LM-cut} & {$h^{\text{add}}$}
  & {LM-cut} & {$h^{\text{add}}$}
  \\
  \hline
  \hline
triangle-tire-4
  & \makecell{30/30 \\ (23.37 $\pm$ 0.66)} 
  & \makecell{30/30 \\ (23.37 $\pm$ 0.66)} 
  & \makecell{30/30 \\ (23.37 $\pm$ 0.66)} 
  & \makecell{30/30 \\ (23.17 $\pm$ 0.69)} 
  & \makecell{30/30 \\ (23.17 $\pm$ 0.69)} 
  & \makecell{30/30 \\ (23.83 $\pm$ 0.71)} 
  & \makecell{30/30 \\ (24.10 $\pm$ 0.76)} 
  & \makecell{30/30 \\ (23.23 $\pm$ 0.76)} 
  & \makecell{30/30 \\ (23.33 $\pm$ 0.80)} 
  \\
triangle-tire-5
  & \makecell{30/30 \\ (28.87 $\pm$ 0.81)} 
  & \makecell{30/30 \\ (28.87 $\pm$ 0.81)} 
  & \makecell{30/30 \\ (28.87 $\pm$ 0.81)} 
  & - 
  & - 
  & \makecell{30/30 \\ (29.27 $\pm$ 0.75)} 
  & \makecell{30/30 \\ (30.23 $\pm$ 0.63)} 
  & \makecell{30/30 \\ (29.43 $\pm$ 0.85)} 
  & \makecell{30/30 \\ (29.83 $\pm$ 0.79)} 
  \\
triangle-tire-6
  & \makecell{30/30 \\ (34.87 $\pm$ 0.94)} 
  & \makecell{30/30 \\ (34.87 $\pm$ 0.94)} 
  & \makecell{30/30 \\ (34.87 $\pm$ 0.94)} 
  & - 
  & - 
  & - 
  & - 
  & \makecell{30/30 \\ (35.70 $\pm$ 0.96)} 
  & \makecell{30/30 \\ (36.90 $\pm$ 0.87)} 
  \\
triangle-tire-7
  & \makecell{30/30 \\ (40.77 $\pm$ 0.91)} 
  & \makecell{30/30 \\ (40.77 $\pm$ 0.91)} 
  & \makecell{30/30 \\ (40.77 $\pm$ 0.91)} 
  & - 
  & - 
  & - 
  & - 
  & \makecell{30/30 \\ (41.43 $\pm$ 0.86)} 
  & \makecell{30/30 \\ (44.33 $\pm$ 1.05)} 
  \\
triangle-tire-8
  & \makecell{30/30 \\ (46.83 $\pm$ 1.12)} 
  & \makecell{30/30 \\ (46.83 $\pm$ 1.12)} 
  & \makecell{30/30 \\ (46.83 $\pm$ 1.12)} 
  & - 
  & - 
  & - 
  & - 
  & \makecell{7/30 \\ (48.00 $\pm$ 3.66)} 
  & \makecell{26/30 \\ (50.77 $\pm$ 1.05)} 
  \\
triangle-tire-10
  & \makecell{30/30 \\ (59.00 $\pm$ 1.11)} 
  & \makecell{30/30 \\ (59.00 $\pm$ 1.11)} 
  & \makecell{30/30 \\ (59.00 $\pm$ 1.11)} 
  & - 
  & - 
  & - 
  & - 
  & \makecell{1/30 \\ (60.00)} 
  & - 
  \\
triangle-tire-9
  & \makecell{30/30 \\ (52.93 $\pm$ 1.27)} 
  & \makecell{30/30 \\ (52.93 $\pm$ 1.27)} 
  & \makecell{30/30 \\ (52.93 $\pm$ 1.27)} 
  & - 
  & - 
  & - 
  & - 
  & \makecell{1/30 \\ (54.00)} 
  & \makecell{1/30 \\ (71.00)} 
  \\
triangle-tire-11
  & \makecell{30/30 \\ (64.77 $\pm$ 1.08)} 
  & \makecell{30/30 \\ (64.77 $\pm$ 1.08)} 
  & \makecell{30/30 \\ (64.77 $\pm$ 1.08)} 
  & - 
  & - 
  & - 
  & - 
  & - 
  & - 
  \\
triangle-tire-12
  & \makecell{30/30 \\ (71.07 $\pm$ 1.21)} 
  & \makecell{30/30 \\ (71.07 $\pm$ 1.21)} 
  & \makecell{30/30 \\ (71.07 $\pm$ 1.21)} 
  & - 
  & - 
  & - 
  & - 
  & - 
  & - 
  \\
triangle-tire-13
  & \makecell{30/30 \\ (76.90 $\pm$ 1.21)} 
  & \makecell{30/30 \\ (76.90 $\pm$ 1.21)} 
  & \makecell{30/30 \\ (76.90 $\pm$ 1.21)} 
  & - 
  & - 
  & - 
  & - 
  & - 
  & - 
  \\
triangle-tire-14
  & \makecell{30/30 \\ (82.80 $\pm$ 1.35)} 
  & \makecell{30/30 \\ (82.80 $\pm$ 1.35)} 
  & \makecell{30/30 \\ (82.80 $\pm$ 1.35)} 
  & - 
  & - 
  & - 
  & - 
  & - 
  & - 
  \\
triangle-tire-15
  & \makecell{30/30 \\ (88.67 $\pm$ 1.37)} 
  & \makecell{30/30 \\ (88.67 $\pm$ 1.37)} 
  & \makecell{30/30 \\ (88.67 $\pm$ 1.37)} 
  & - 
  & - 
  & - 
  & - 
  & - 
  & - 
  \\
triangle-tire-16
  & \makecell{30/30 \\ (94.83 $\pm$ 1.29)} 
  & \makecell{30/30 \\ (94.83 $\pm$ 1.29)} 
  & \makecell{30/30 \\ (94.83 $\pm$ 1.29)} 
  & - 
  & - 
  & - 
  & - 
  & - 
  & - 
  \\
triangle-tire-17
  & \makecell{30/30 \\ (100.80 $\pm$ 1.21)} 
  & \makecell{30/30 \\ (100.80 $\pm$ 1.21)} 
  & \makecell{30/30 \\ (100.80 $\pm$ 1.21)} 
  & - 
  & - 
  & - 
  & - 
  & - 
  & - 
  \\
triangle-tire-18
  & \makecell{30/30 \\ (106.50 $\pm$ 1.44)} 
  & \makecell{30/30 \\ (106.50 $\pm$ 1.44)} 
  & \makecell{30/30 \\ (106.50 $\pm$ 1.44)} 
  & - 
  & - 
  & - 
  & - 
  & - 
  & - 
  \\
triangle-tire-19
  & \makecell{30/30 \\ (112.50 $\pm$ 1.56)} 
  & \makecell{30/30 \\ (112.50 $\pm$ 1.56)} 
  & \makecell{30/30 \\ (112.50 $\pm$ 1.56)} 
  & - 
  & - 
  & - 
  & - 
  & - 
  & - 
  \\
triangle-tire-20
  & \makecell{30/30 \\ (118.43 $\pm$ 1.48)} 
  & \makecell{30/30 \\ (118.43 $\pm$ 1.48)} 
  & \makecell{30/30 \\ (118.43 $\pm$ 1.48)} 
  & - 
  & - 
  & - 
  & - 
  & - 
  & - 
  \\
  \hline
\end{tabular}
\end{scriptsize}
\end{center}
\caption{Coverage (number of successful trials to reach the goal) for a
  selection of problems and planners. Mean cost to reach the goal and 95\% CI
  for cost is given in brackets.}
\label{tab:prob-res-ttw}
\end{sidewaystable*}

\begin{sidewaystable*}
\begin{center}
\begin{scriptsize}
\begin{tabular}{|l|ccc|cc|cc|cc|}
\hline
\multirow{2}{*}{Problem}
  & \multicolumn{3}{c|}{ASNet}
  & \multicolumn{2}{c|}{ILAO}
  & \multicolumn{2}{c|}{LRTDP}
  & \multicolumn{2}{c|}{SSiPP}
  \\
  & {LM-cut} & {$h^{\text{add}}$} & {$h^{\text{add}}$, no LM}
  & {LM-cut} & {$h^{\text{add}}$}
  & {LM-cut} & {$h^{\text{add}}$}
  & {LM-cut} & {$h^{\text{add}}$}
  \\
  \hline
  \hline
prob-bw-n9-s1
  & - 
  & \makecell{30/30 \\ (26.37 $\pm$ 1.60)} 
  & \makecell{30/30 \\ (26.37 $\pm$ 1.60)} 
  & \makecell{30/30 \\ (27.03 $\pm$ 1.75)} 
  & \makecell{30/30 \\ (26.83 $\pm$ 1.83)} 
  & \makecell{3/30 \\ (24.67 $\pm$ 3.79)} 
  & \makecell{30/30 \\ (27.00 $\pm$ 1.81)} 
  & \makecell{30/30 \\ (184.77 $\pm$ 80.73)} 
  & \makecell{30/30 \\ (25.53 $\pm$ 1.58)} 
  \\
prob-bw-n9-s2
  & - 
  & \makecell{30/30 \\ (32.57 $\pm$ 1.79)} 
  & \makecell{30/30 \\ (32.57 $\pm$ 1.79)} 
  & \makecell{30/30 \\ (31.43 $\pm$ 2.21)} 
  & \makecell{30/30 \\ (32.67 $\pm$ 2.01)} 
  & \makecell{30/30 \\ (18.80 $\pm$ 1.81)} 
  & \makecell{30/30 \\ (33.17 $\pm$ 1.70)} 
  & \makecell{13/30 \\ (480.08 $\pm$ 147.47)} 
  & \makecell{30/30 \\ (32.87 $\pm$ 1.91)} 
  \\
prob-bw-n9-s3
  & - 
  & \makecell{30/30 \\ (19.27 $\pm$ 1.99)} 
  & \makecell{30/30 \\ (19.27 $\pm$ 1.99)} 
  & \makecell{30/30 \\ (21.03 $\pm$ 1.78)} 
  & \makecell{30/30 \\ (21.67 $\pm$ 2.01)} 
  & - 
  & \makecell{30/30 \\ (20.47 $\pm$ 1.65)} 
  & \makecell{30/30 \\ (27.53 $\pm$ 9.11)} 
  & \makecell{30/30 \\ (20.33 $\pm$ 1.72)} 
  \\
  \hline
prob-bw-n10-s1
  & - 
  & \makecell{30/30 \\ (24.60 $\pm$ 1.97)} 
  & \makecell{30/30 \\ (24.60 $\pm$ 1.97)} 
  & \makecell{30/30 \\ (25.50 $\pm$ 1.90)} 
  & \makecell{30/30 \\ (26.87 $\pm$ 1.71)} 
  & - 
  & \makecell{30/30 \\ (24.03 $\pm$ 1.23)} 
  & \makecell{30/30 \\ (78.47 $\pm$ 24.76)} 
  & \makecell{30/30 \\ (25.03 $\pm$ 1.57)} 
  \\
prob-bw-n10-s2
  & - 
  & \makecell{30/30 \\ (33.87 $\pm$ 2.01)} 
  & \makecell{30/30 \\ (33.87 $\pm$ 2.01)} 
  & - 
  & \makecell{30/30 \\ (36.37 $\pm$ 1.96)} 
  & - 
  & \makecell{30/30 \\ (34.27 $\pm$ 1.64)} 
  & \makecell{14/30 \\ (484.50 $\pm$ 173.12)} 
  & \makecell{30/30 \\ (35.27 $\pm$ 1.79)} 
  \\
prob-bw-n10-s3
  & - 
  & \makecell{30/30 \\ (28.23 $\pm$ 1.96)} 
  & \makecell{30/30 \\ (28.73 $\pm$ 2.17)} 
  & - 
  & \makecell{30/30 \\ (29.90 $\pm$ 2.01)} 
  & - 
  & \makecell{30/30 \\ (28.13 $\pm$ 1.77)} 
  & \makecell{30/30 \\ (127.20 $\pm$ 33.51)} 
  & \makecell{30/30 \\ (28.60 $\pm$ 1.71)} 
  \\
  \hline
prob-bw-n15-s1
  & - 
  & \makecell{30/30 \\ (46.77 $\pm$ 2.52)} 
  & \makecell{30/30 \\ (49.23 $\pm$ 2.35)} 
  & - 
  & \makecell{30/30 \\ (48.87 $\pm$ 2.83)} 
  & - 
  & \makecell{30/30 \\ (50.10 $\pm$ 1.92)} 
  & \makecell{30/30 \\ (94.23 $\pm$ 10.12)} 
  & \makecell{30/30 \\ (51.27 $\pm$ 1.47)} 
  \\
prob-bw-n15-s2
  & - 
  & \makecell{30/30 \\ (55.23 $\pm$ 2.31)} 
  & \makecell{30/30 \\ (55.50 $\pm$ 2.45)} 
  & - 
  & \makecell{30/30 \\ (57.67 $\pm$ 2.63)} 
  & - 
  & \makecell{30/30 \\ (57.10 $\pm$ 2.49)} 
  & \makecell{30/30 \\ (185.00 $\pm$ 33.55)} 
  & \makecell{30/30 \\ (58.60 $\pm$ 2.00)} 
  \\
prob-bw-n15-s3
  & - 
  & \makecell{30/30 \\ (46.53 $\pm$ 2.60)} 
  & \makecell{30/30 \\ (48.50 $\pm$ 2.33)} 
  & - 
  & \makecell{30/30 \\ (46.40 $\pm$ 2.49)} 
  & - 
  & \makecell{30/30 \\ (45.13 $\pm$ 1.83)} 
  & \makecell{30/30 \\ (249.20 $\pm$ 50.41)} 
  & \makecell{30/30 \\ (46.00 $\pm$ 2.07)} 
  \\
  \hline
prob-bw-n20-s1
  & - 
  & \makecell{30/30 \\ (65.93 $\pm$ 2.39)} 
  & \makecell{30/30 \\ (70.33 $\pm$ 2.51)} 
  & - 
  & \makecell{30/30 \\ (69.63 $\pm$ 2.54)} 
  & - 
  & \makecell{30/30 \\ (70.70 $\pm$ 3.36)} 
  & - 
  & \makecell{30/30 \\ (70.00 $\pm$ 2.87)} 
  \\
prob-bw-n20-s2
  & - 
  & \makecell{30/30 \\ (76.77 $\pm$ 2.11)} 
  & \makecell{30/30 \\ (76.77 $\pm$ 2.11)} 
  & - 
  & \makecell{30/30 \\ (73.87 $\pm$ 2.17)} 
  & - 
  & \makecell{30/30 \\ (79.10 $\pm$ 2.73)} 
  & - 
  & \makecell{30/30 \\ (83.53 $\pm$ 3.16)} 
  \\
prob-bw-n20-s3
  & - 
  & \makecell{30/30 \\ (69.53 $\pm$ 2.81)} 
  & \makecell{30/30 \\ (77.30 $\pm$ 2.60)} 
  & - 
  & \makecell{30/30 \\ (74.60 $\pm$ 2.82)} 
  & - 
  & \makecell{30/30 \\ (76.27 $\pm$ 3.44)} 
  & - 
  & \makecell{30/30 \\ (78.20 $\pm$ 3.29)} 
  \\
  \hline
prob-bw-n25-s1
  & - 
  & \makecell{30/30 \\ (98.27 $\pm$ 2.99)} 
  & \makecell{30/30 \\ (99.47 $\pm$ 2.89)} 
  & - 
  & - 
  & - 
  & \makecell{17/30 \\ (100.94 $\pm$ 4.60)} 
  & - 
  & \makecell{28/30 \\ (323.96 $\pm$ 91.46)} 
  \\
prob-bw-n25-s2
  & - 
  & \makecell{30/30 \\ (91.50 $\pm$ 2.47)} 
  & \makecell{30/30 \\ (91.77 $\pm$ 2.64)} 
  & - 
  & - 
  & - 
  & \makecell{27/30 \\ (100.78 $\pm$ 3.16)} 
  & - 
  & \makecell{30/30 \\ (145.63 $\pm$ 27.99)} 
  \\
prob-bw-n25-s3
  & - 
  & \makecell{30/30 \\ (89.70 $\pm$ 2.59)} 
  & \makecell{30/30 \\ (85.90 $\pm$ 2.00)} 
  & - 
  & - 
  & - 
  & \makecell{15/30 \\ (95.73 $\pm$ 5.99)} 
  & - 
  & \makecell{29/30 \\ (163.41 $\pm$ 35.31)} 
  \\
  \hline
prob-bw-n30-s1
  & - 
  & \makecell{30/30 \\ (116.43 $\pm$ 3.01)} 
  & \makecell{30/30 \\ (117.23 $\pm$ 2.77)} 
  & - 
  & - 
  & - 
  & \makecell{2/30 \\ (107.50 $\pm$ 44.47)} 
  & - 
  & \makecell{27/30 \\ (340.37 $\pm$ 63.31)} 
  \\
prob-bw-n30-s2
  & - 
  & \makecell{30/30 \\ (111.20 $\pm$ 3.36)} 
  & \makecell{30/30 \\ (113.27 $\pm$ 3.56)} 
  & - 
  & - 
  & - 
  & - 
  & - 
  & \makecell{21/30 \\ (418.38 $\pm$ 82.95)} 
  \\
prob-bw-n30-s3
  & - 
  & \makecell{30/30 \\ (117.30 $\pm$ 3.33)} 
  & \makecell{30/30 \\ (119.00 $\pm$ 2.88)} 
  & - 
  & - 
  & - 
  & - 
  & - 
  & \makecell{16/30 \\ (373.31 $\pm$ 83.93)} 
  \\
  \hline
prob-bw-n35-s1
  & - 
  & \makecell{30/30 \\ (138.80 $\pm$ 3.37)} 
  & \makecell{30/30 \\ (138.87 $\pm$ 3.04)} 
  & - 
  & - 
  & - 
  & - 
  & - 
  & \makecell{1/30 \\ (366.00)} 
  \\
prob-bw-n35-s2
  & - 
  & \makecell{30/30 \\ (137.00 $\pm$ 3.12)} 
  & \makecell{30/30 \\ (137.70 $\pm$ 3.41)} 
  & - 
  & - 
  & - 
  & - 
  & - 
  & \makecell{3/30 \\ (283.67 $\pm$ 199.76)} 
  \\
prob-bw-n35-s3
  & - 
  & \makecell{30/30 \\ (139.27 $\pm$ 3.31)} 
  & \makecell{30/30 \\ (139.33 $\pm$ 3.62)} 
  & - 
  & - 
  & - 
  & - 
  & - 
  & \makecell{6/30 \\ (287.33 $\pm$ 137.80)} 
  \\
  \hline
\end{tabular}
\end{scriptsize}
\end{center}
\caption{\cref{tab:prob-res-ttw} repeated for the Probabilistic Blocks World domain.}
\label{tab:prob-res-pbw}
\end{sidewaystable*}

\begin{sidewaystable*}
\begin{center}
\begin{scriptsize}
\begin{tabular}{|l|ccc|cc|cc|cc|}
\hline
\multirow{2}{*}{Problem}
  & \multicolumn{3}{c|}{ASNet}
  & \multicolumn{2}{c|}{ILAO}
  & \multicolumn{2}{c|}{LRTDP}
  & \multicolumn{2}{c|}{SSiPP}
  \\
  & {LM-cut} & {$h^{\text{add}}$} & {$h^{\text{add}}$, no LM}
  & {LM-cut} & {$h^{\text{add}}$}
  & {LM-cut} & {$h^{\text{add}}$}
  & {LM-cut} & {$h^{\text{add}}$}
  \\
  \hline
  \hline
  cosanostra-n10
  & \makecell{30/30 \\ (34.00 $\pm$ 0)} 
  & \makecell{30/30 \\ (34.00 $\pm$ 0)} 
  & - 
  & \makecell{30/30 \\ (34.00 $\pm$ 0)} 
  & \makecell{30/30 \\ (34.00 $\pm$ 0)} 
  & \makecell{30/30 \\ (34.00 $\pm$ 0)} 
  & \makecell{30/30 \\ (34.00 $\pm$ 0)} 
  & \makecell{30/30 \\ (34.00 $\pm$ 0)} 
  & \makecell{30/30 \\ (34.00 $\pm$ 0)} 
  \\
cosanostra-n11
  & \makecell{30/30 \\ (37.00 $\pm$ 0)} 
  & \makecell{30/30 \\ (37.00 $\pm$ 0)} 
  & - 
  & \makecell{30/30 \\ (37.00 $\pm$ 0)} 
  & \makecell{30/30 \\ (37.00 $\pm$ 0)} 
  & \makecell{30/30 \\ (37.00 $\pm$ 0)} 
  & \makecell{30/30 \\ (37.00 $\pm$ 0)} 
  & \makecell{30/30 \\ (37.00 $\pm$ 0)} 
  & \makecell{30/30 \\ (37.00 $\pm$ 0)} 
  \\
cosanostra-n12
  & \makecell{30/30 \\ (40.00 $\pm$ 0)} 
  & \makecell{30/30 \\ (40.00 $\pm$ 0)} 
  & - 
  & \makecell{30/30 \\ (40.00 $\pm$ 0)} 
  & \makecell{30/30 \\ (40.00 $\pm$ 0)} 
  & \makecell{30/30 \\ (40.00 $\pm$ 0)} 
  & \makecell{30/30 \\ (40.00 $\pm$ 0)} 
  & \makecell{30/30 \\ (40.00 $\pm$ 0)} 
  & \makecell{30/30 \\ (40.00 $\pm$ 0)} 
  \\
cosanostra-n13
  & \makecell{30/30 \\ (43.00 $\pm$ 0)} 
  & \makecell{30/30 \\ (43.00 $\pm$ 0)} 
  & - 
  & \makecell{30/30 \\ (43.00 $\pm$ 0)} 
  & \makecell{30/30 \\ (43.00 $\pm$ 0)} 
  & - 
  & \makecell{30/30 \\ (43.00 $\pm$ 0)} 
  & - 
  & \makecell{30/30 \\ (43.00 $\pm$ 0)} 
  \\
cosanostra-n14
  & \makecell{30/30 \\ (46.00 $\pm$ 0)} 
  & \makecell{30/30 \\ (46.00 $\pm$ 0)} 
  & - 
  & - 
  & \makecell{30/30 \\ (46.00 $\pm$ 0)} 
  & - 
  & - 
  & - 
  & - 
  \\
cosanostra-n15
  & \makecell{30/30 \\ (49.00 $\pm$ 0)} 
  & \makecell{30/30 \\ (49.00 $\pm$ 0)} 
  & - 
  & - 
  & \makecell{30/30 \\ (49.00 $\pm$ 0)} 
  & - 
  & - 
  & - 
  & - 
  \\
cosanostra-n20
  & \makecell{30/30 \\ (64.00 $\pm$ 0)} 
  & \makecell{30/30 \\ (64.00 $\pm$ 0)} 
  & - 
  & - 
  & - 
  & - 
  & - 
  & - 
  & - 
  \\
  \hline
\end{tabular}
\end{scriptsize}
\end{center}
\caption{\cref{tab:prob-res-ttw} repeated for the CosaNostra Pizza domain.}
\label{tab:prob-res-cn}
\end{sidewaystable*} 
 \end{document}